\documentclass[runningheads]{llncs}
\usepackage{amsmath}
\usepackage{esvect}
\usepackage{url}
\usepackage{algorithm}        
\usepackage{algpseudocode}    
\usepackage{multirow}
\usepackage{graphicx} 
\usepackage{booktabs}
\usepackage{amssymb}
\usepackage[T1]{fontenc}
\usepackage{graphicx}
\usepackage{enumitem}
\setlist{leftmargin=*,itemsep=1pt,topsep=2pt,parsep=0pt}
\begin{document}
\title{CAViT - Channel-Aware Vision Transformer for Dynamic Feature Fusion}
%
%
\author{Aon Safdar\orcidID{0000-0003-3039-1017} \and
Mohamed Saadeldin\orcidID{0000-0003-4104-9524}}
\authorrunning{A. Safdar and M. Saadeldin}
%
\institute{School of Computer Science, University College Dublin, Republic of Ireland
\email{aon.safdar@ucdconnect.ie, mohamed.saadeldin@ucd.ie}}
\maketitle              
\begin{abstract}
Vision Transformers (ViTs) have demonstrated strong performance across a range of computer vision tasks by modeling long-range spatial interactions via self-attention. However, channel-wise mixing in ViTs remains static, relying on fixed multilayer perceptrons (MLPs) that lack adaptability to input content. We introduce \textbf{CAViT}, a dual-attention architecture that replaces the static MLP with a dynamic, attention-based mechanism for feature interaction. Each Transformer block in CAViT \footnote{Code available at \url{https://github.com/aonsafdar/CAVit}} performs spatial self-attention followed by channel-wise self-attention, allowing the model to dynamically recalibrate feature representations based on global image context. This unified and content-aware token mixing strategy enhances representational expressiveness without increasing depth or complexity. We validate CAViT across five benchmark datasets spanning both natural and medical domains, where it outperforms the standard ViT$_\text{tiny}$ baseline by up to \textbf{+3.6\%} in accuracy, while reducing parameter count and FLOPs by over \textbf{30\%}. Qualitative attention maps reveal sharper and semantically meaningful activation patterns, validating the effectiveness of our attention-driven token mixing.

\end{abstract}
%
%
%
%
\section{Introduction}
\label{sec:introduction}

\label{sec:intro}

\begin{figure}[t]
    \centering
    \includegraphics[width=0.95\linewidth]{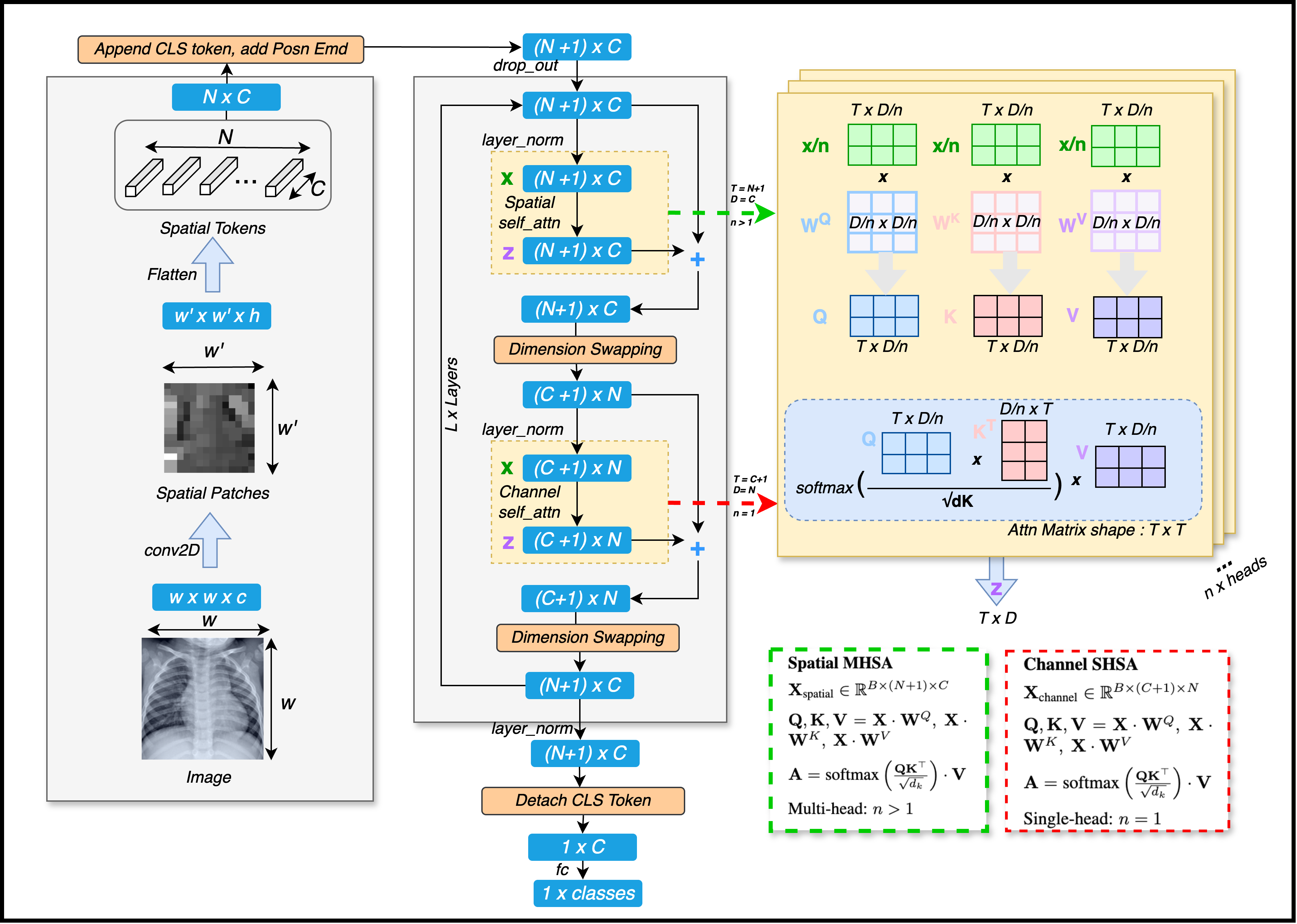}
    \caption{\textbf{CAViT overview.} (Left) Standard tokenization. (Middle) Each block applies spatial MHSA on tokens
$\mathbf{x}\!\in\!\mathbb{R}^{B\times(N{+}1)\times C}$, then swaps dimensions to
$\mathbb{R}^{B\times(C{+}1)\times N}$ to treat channels as tokens and applies
single-head channel self-attention (SHSA); the swap is then reversed. For spatial
MHSA: $T\!=\!N{+}1$, $D\!=\!C$, $n{>}1$; for channel SHSA: $T\!=\!C{+}1$, $D\!=\!N$, $n\!=\!1$.
(Right) Standard $Q,K,V$ projections and softmax over $T\times T$.}
    \label{fig:overview}
\end{figure}

Vision Transformers (ViTs) have reshaped modern computer vision by replacing convolutions with self-attention to model long-range dependencies~\cite{newdosovitskiy2021an,touvron_training_2021}. While their flexibility and scalability have led to impressive results across classification, detection, and segmentation \cite{han_survey_2023}, ViTs retain an architectural asymmetry: spatial mixing is dynamic and data-dependent via Multi-Head Self-Attention (MHSA), but channel mixing is static, performed through fixed MLP layers. These MLPs linearly combine features across channels without awareness of image content, potentially limiting the model’s ability to adapt to different visual structures.

This static channel mixing contrasts with the semantic role of channels in vision, where feature dimensions often represent meaningful filters (e.g., textures, parts, or semantic cues). Several works have attempted to address this issue by introducing channel attention mechanisms—such as SE-Net~\cite{Hu_2018_CVPR} and Cross-Covariance Attention~\cite{ali_xcit_2021}—or dynamic MLPs~\cite{liu_pay_2021}, but these often require additional modules or incur structural overhead.

We propose a simple and general solution by replacing the static MLP with a second self-attention stage that operates across feature channels. By transposing the input tensor and treating channels as tokens (Figure \ref{fig:overview}), our method enables ViTs to learn inter-channel dependencies conditioned on global image context dynamically. This dual-attention design is seamlessly integrated into the standard Transformer block, improving both representational power and interpretability without increasing model depth or adding new modules.

\vspace{0.5 em}
\noindent
\textbf{Our key contributions are:}
\begin{itemize}
    \item We propose \textbf{CAViT}, a simple and generic architectural modification that replaces static MLPs in ViT blocks with channel-wise attention via dimension swapping, enabling adaptive inter-channel interactions.
    
    \item Our dual-attention formulation unifies spatial and channel-wise token mixing under a fully attention-driven framework, improving expressivity while preserving simplicity.
    
    \item We provide a theoretical and architectural motivation for dynamic channel mixing, showing how attention in the channel domain enhances information flow across feature dimensions and improves feature alignment.
    
    \item Through experiments on five diverse datasets across natural and medical imaging, we demonstrate consistent improvements in accuracy and efficiency over standard ViT baselines, both quantitatively and qualitatively.
\end{itemize}

\section{Related Work}
\label{sec:RelatedWork}
\textbf{Vision Transformers.}
Transformers, initially developed for NLP~\cite{vaswani_attention_2023}, were adapted to vision with ViT~\cite{newdosovitskiy2021an}, which uses tokenized image patches and MHSA. Subsequent variants such as DeiT~\cite{touvron_training_2021} and Swin Transformer~\cite{liu_swin_2021} improved training efficiency and scalability. Token-based architectures now serve as the backbone of many vision foundation models~\cite{bordes_introduction_2024,zhou_image_2024}, motivating exploration of richer token and feature mixing strategies~\cite{khan_transformers_2022,shamshad_transformers_2023}.

\noindent\textbf{Information Mixing Strategies.}
While standard ViTs mix spatial and channel information using MHSA and static MLPs, respectively, recent work has proposed hybrid designs~\cite{wu_cvt_2021,kahatapitiya_swat_2023,zheng_lightweight_2023} or exclusive MLP/convolutional mixers~\cite{tolstikhin_mlp-mixer_2021,trockman_patches_2022}. Our work follows this line but investigates replacing the MLP with attention-based channel mixing to enhance context-aware feature interaction.

\noindent\textbf{Channel Attention in Vision.}
SE-Net~\cite{Hu_2018_CVPR}, CBAM~\cite{Woo_2018_ECCV}, and Dual Attention~\cite{Fu_2019_CVPR} demonstrate that modeling channel dependencies improves CNN performance. These techniques, however, rely on static operations. We build on their insights by incorporating learnable, dynamic channel attention directly into ViT blocks.

\noindent\textbf{Dynamic Channel Mixing in Transformers.}
Content-adaptive strategies such as gMLP~\cite{liu_pay_2021}, XCiT~\cite{ali_xcit_2021}, and DaViT~\cite{ding_davit_2022} show that dynamic channel interactions improve expressiveness. Ablation studies in~\cite{yuan_tokens--token_2021} suggest that even lightweight channel attention can benefit ViTs. Unlike prior works that introduced channel attention as auxiliary components or external modules, our approach explores replacing the static MLP within the core Transformer block, enabling a unified architecture where both spatial and channel interactions are handled exclusively via self-attention, thus preserving the Transformer’s structural simplicity while improving its representational expressiveness.

\section{CAViT: Channel-Aware Vision Transformer with Dynamic Channel Mixing}
\label{sec:proposed}

Our proposed architecture, CAViT, enhances the standard Vision Transformer (ViT) by introducing a dual-attention mechanism within each Transformer block. As illustrated in ure~\ref{fig:arch}, the core idea is to decouple the spatial and channel-wise mixing stages and treat both as attention operations. This is achieved through a simple yet effective architectural modification: replacing the MLP-based feedforward layer with a second attention stage applied across feature channels via dimension swapping.

\subsection{Architecture Overview}
As highlighted in Figure~\ref{fig:overview}, the conventional ViT architecture processes an input image $\mathbf{I} \in \mathbb{R}^{W \times W \times 3}$ by dividing it into non-overlapping patches of size $w \times w$ using linear or hard-split convolutional projections where each patch is represented by a vector of dimension $C$. This results in a sequence of $N = \frac{W \times W}{w^2}$ patch tokens, forming an input tensor $\mathbf{x} \in \mathbb{R}^{B \times N \times C}$, where $B$ is the batch size, $N$ is the number of spatial tokens, and $C$ is the embedding dimension. A learnable classification token (CLS) is prepended to this sequence, and positional embeddings are added, yielding $\mathbf{x} \in \mathbb{R}^{B \times (N+1) \times C}$. In CAVit, this sequence undergoes spatial attention followed by dimension swapping and channel attention before reversing the swap and handling CLS token to ensure global aggregation.

\begin{figure}[t]
  \centering
   \includegraphics[width=1.0\linewidth]{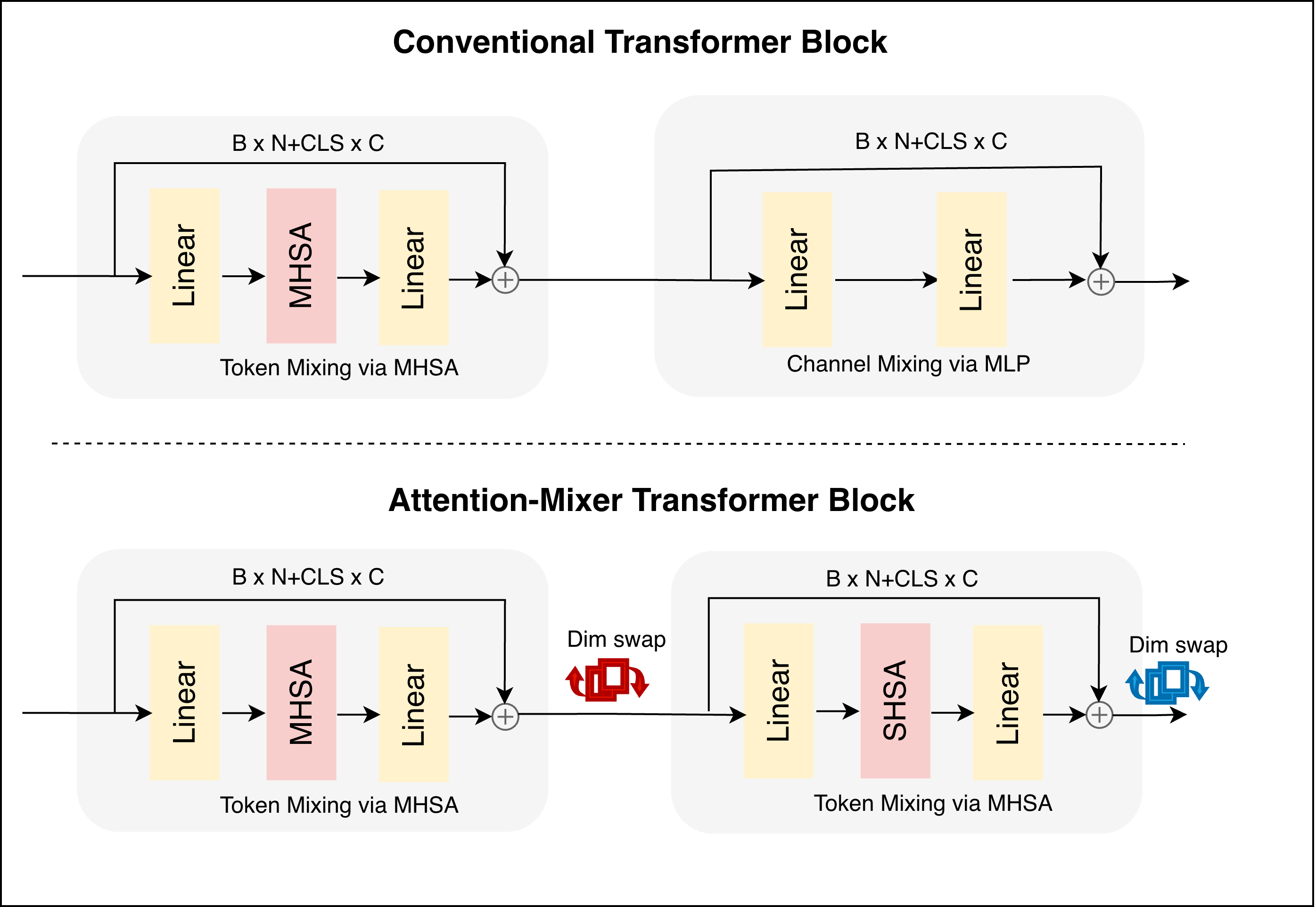}

   \caption{\textbf{CAVit Transformer Block.} We propose a dual-attention Transformer block that models both spatial and channel-wise interactions using two sequential self-attention stages. The first stage applies standard MHSA over spatial tokens and operates across the token axis. The second stage performs channel-wise attention by swapping spatial and channel dimensions and applying SHSA  across channels.}
   \label{fig:arch}
\end{figure}

Conventional ViT block (top of Figure~\ref{fig:arch}) consists of two core stages: (1) a Multi-Head Self-Attention (MHSA) layer that operates across the sequence of spatial tokens, and (2) a feedforward MLP that performs channel-wise mixing for each token independently. This asymmetry in treatment, i.e, adaptive attention across space, but static linear mixing across channels, limits representational flexibility. In CAViT (bottom of Figure~\ref{fig:arch}), we preserve the MHSA spatial attention but replace the MLP with a second attention module, enabling channel-wise mixing via a transposition that allows attention to operate across feature channels. This modification reframes the Transformer block as an attention-only unit, enhancing both spatial and channel interactions through learnable, content-adaptive mechanisms.

\subsection{Dynamic Feature Fusion}
The MLP accounts for a large fraction of parameters and FLOPs in ViT, yet it acts as a token-wise linear transformation with nonlinear gating. This design neither exploits inter-channel relationships nor adapts to content. In contrast, using self-attention across channels introduces a dynamic, data-driven inductive bias, better aligned with how semantic concepts manifest across channel dimensions in deep networks (e.g., texture detectors, part filters). Spatial and channel attention capture complementary cues. While spatial MHSA models long-range positional relationships, channel attention models dependencies among feature activations. Incorporating both into a single, symmetric structure improves model expressiveness.

\begin{figure}[t]
  \centering
   \includegraphics[width=1.0\linewidth]{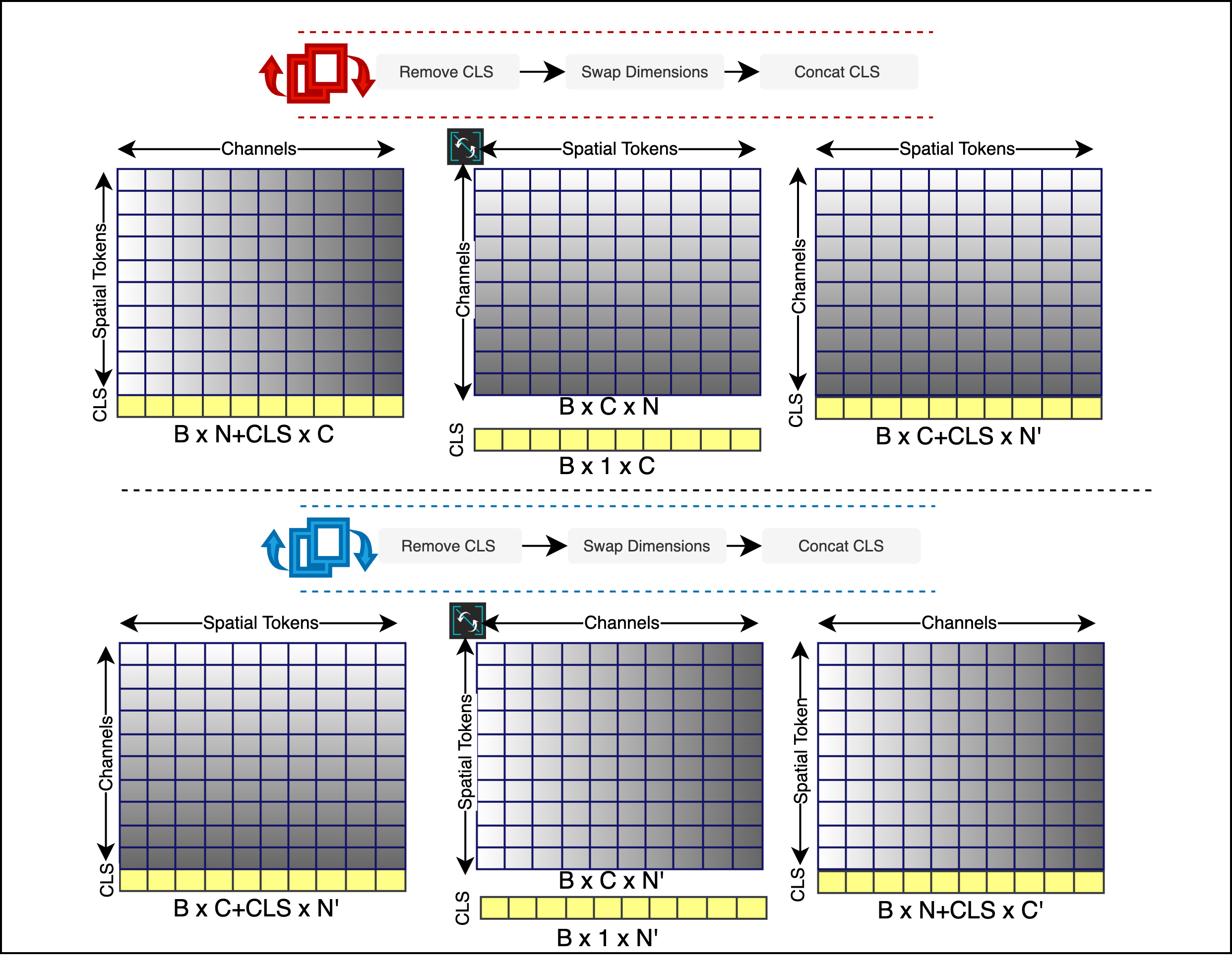}
   \caption{\textbf{Dimension Swapping Mechanism.} Illustration of our dimension swapping logic used to enable channel-wise attention. The class token (CLS) is first separated from the input tensor $\mathbf{x} \in \mathbb{R}^{B \times (N{+}1) \times C}$, where $B$ is batch size, $N$ is the number of spatial tokens, and $C$ is the channel dimension. The remaining spatial tokens are transposed from shape $\mathbb{R}^{B \times N \times C}$ to $\mathbb{R}^{B \times C \times N}$, effectively treating channels as attention tokens. The CLS token is then concatenated back, resulting in a new tensor of shape $\mathbb{R}^{B \times (C{+}1) \times N}$. Single-head self-attention (SHSA) is applied across this transformed token dimension. The reverse operation restores the original layout, enabling standard downstream processing.}
   \label{fig:dimswap}
\end{figure}

\noindent\textbf{CLS Token Handling.}
The CLS token serves as a global context aggregator\footnote{Some pooling-based classification heads (e.g., average or max pooling) avoid using a CLS token altogether. While our method is compatible with such configurations, we focus on standard ViT settings where CLS is retained, to ensure fair comparison and compatibility with vanilla backbones.}. If improperly handled (e.g., swapped alongside spatial tokens (See Table \ref{tab:ablation}),  it loses its global aggregation role and becomes semantically entangled. Our design maintains the integrity of the CLS token (See Figure \ref{fig:dimswap}), allowing it to aggregate attention information during both spatial and channel attention phases.

\noindent\textbf{Dimension Swapping for Channel Attention.}
As illustrated in Figure~\ref{fig:dimswap}, to apply attention across channels, we first transpose the input representation such that the feature dimensions become the sequence axis. Specifically, after the spatial MHSA stage, the tensor $\mathbf{x} \in \mathbb{R}^{B \times (N{+}1) \times C}$ is split into the class token $\mathbf{x}_{\text{cls}} \in \mathbb{R}^{B \times 1 \times C}$ and the spatial tokens $\mathbf{x}_{\text{spat}} \in \mathbb{R}^{B \times N \times C}$. We then transpose the spatial token tensor to $\mathbf{x}_{\text{trans}} \in \mathbb{R}^{B \times C \times N}$, thereby interpreting channels as sequence tokens. To ensure the CLS token participates in global channel-wise interactions, it is explicitly reshaped to $\mathbf{x}_{\text{cls}} \in \mathbb{R}^{B \times 1 \times N}$, ensuring alignment along the sequence axis before concatenation. This results in a tensor of shape $\mathbb{R}^{B \times (C{+}1) \times N}$ for channel-wise attention\footnote{In our experiments, we set $C = N$, avoiding the need for explicit reshaping. For general cases with $C \ne N$, a linear projection of the CLS token ensures compatibility.}.

\noindent\textbf{Single-Head Channel Attention}
Attention is now applied across the channel-token sequence. Unlike the spatial attention stage, we use a single-head self-attention (SHSA) mechanism here. The rationale is that in the transposed space, each channel token already encodes the global image context, and dividing attention across multiple heads may dilute this abstraction. While SHSA introduces a potential computational bottleneck, it avoids over-fragmentation in the image space and preserves full spatial context within each channel token, enabling coherent and content-adaptive mixing across channels. The effects are further verified in Section~\ref{subsec:ablation}. After attention, we reverse the dimension swap: the CLS token is again extracted, the remaining sequence is transposed back to shape $\mathbb{R}^{B \times N \times C}$, and the CLS token is prepended. The next Transformer block thus receives a representation enriched with both spatial and channel-level interactions.

\section{Experiments And Results}
\label{sec:experiments}
\subsection{Datasets and Metrics}
We evaluate our method on five classification datasets spanning both natural and medical imaging domains: CIFAR-10~\cite{cifar10}, Malaria~\cite{malaria_dataset}, Cats vs Dogs~\cite{dogs-vs-cats}, PneumoniaMNIST, and BreastMNIST~\cite{yang_medmnist_2023}. These datasets (“See \ref{tab:dataset-summary} for an overview.”) encompass binary and multi-class settings and cover diverse imaging modalities including RGB photographs, microscopy, X-ray, and ultrasound. To ensure architectural consistency across datasets and facilitate fair comparison, all input images are resized to $224 \times 224$ pixels.

We compare ViT$_{\text{tiny}}$ and CAViT$_{\text{tiny}}$ in terms of performance and efficiency. For each model, we report the best Top-1 test accuracy achieved over 100 training epochs. To evaluate model complexity, we use two widely adopted and system-independent metrics: parameter count (total number of learnable weights) and GFLOPs (floating-point operations per forward pass). These metrics provide reproducible estimates of memory footprint and compute cost, respectively.
\begin{table}[t]
\centering
\caption{\textbf{Overview of datasets used for evaluation.} We benchmark ViT$_{\text{tiny}}$ and CAViT$_{\text{tiny}}$ variants across natural and medical image classification datasets, spanning binary (BC) and multi-class (MC) tasks with diverse modalities. All images are resized to $224 \times 224$ resolution.}
\vspace{0.5em}

\label{tab:dataset-summary}
\resizebox{\linewidth}{!}{
\begin{tabular}{lccc}
\toprule
\textbf{Name} & \textbf{Modality} & \textbf{Task Type} & \textbf{\# Samples} \\
\midrule
CIFAR-10~\cite{cifar10}         & RGB (Natural Scene)       & MC (10)    & 60,000 \\
Cats vs Dogs~\cite{dogs-vs-cats}                   & RGB (Natural Scene)       & BC (2)    & 25,000 \\
Malaria~\cite{malaria_dataset}                        & Microscopy (Blood Smear)  & BC (2)    & 27,588 \\
PneumoniaMNIST~\cite{yang_medmnist_2023} & Chest X-ray                 & BC (2)    & 5,856 \\
BreastMNIST~\cite{yang_medmnist_2023}    & Ultrasound (Breast)        & BC (2)    & 780 \\
\bottomrule
\end{tabular}
}
\end{table}
We implement both models in PyTorch using the \texttt{timm}~\cite{Wightman_PyTorch_Image_Models} components. We train with SGD optimizer, an initial learning rate of 0.001, and no aggressive augmentation or regularization, to isolate the effects of our architectural modifications. All experiments are conducted on a Linux machine with an NVIDIA RTX 4090 GPU (CUDA 12.4), PyTorch 2.6.0, and Python 3.10. For reproducibility, we fix the random seed to 42.

\subsection{Results and Analysis}
\begin{figure}[t]
  \fbox{%
    \includegraphics[width=\linewidth]{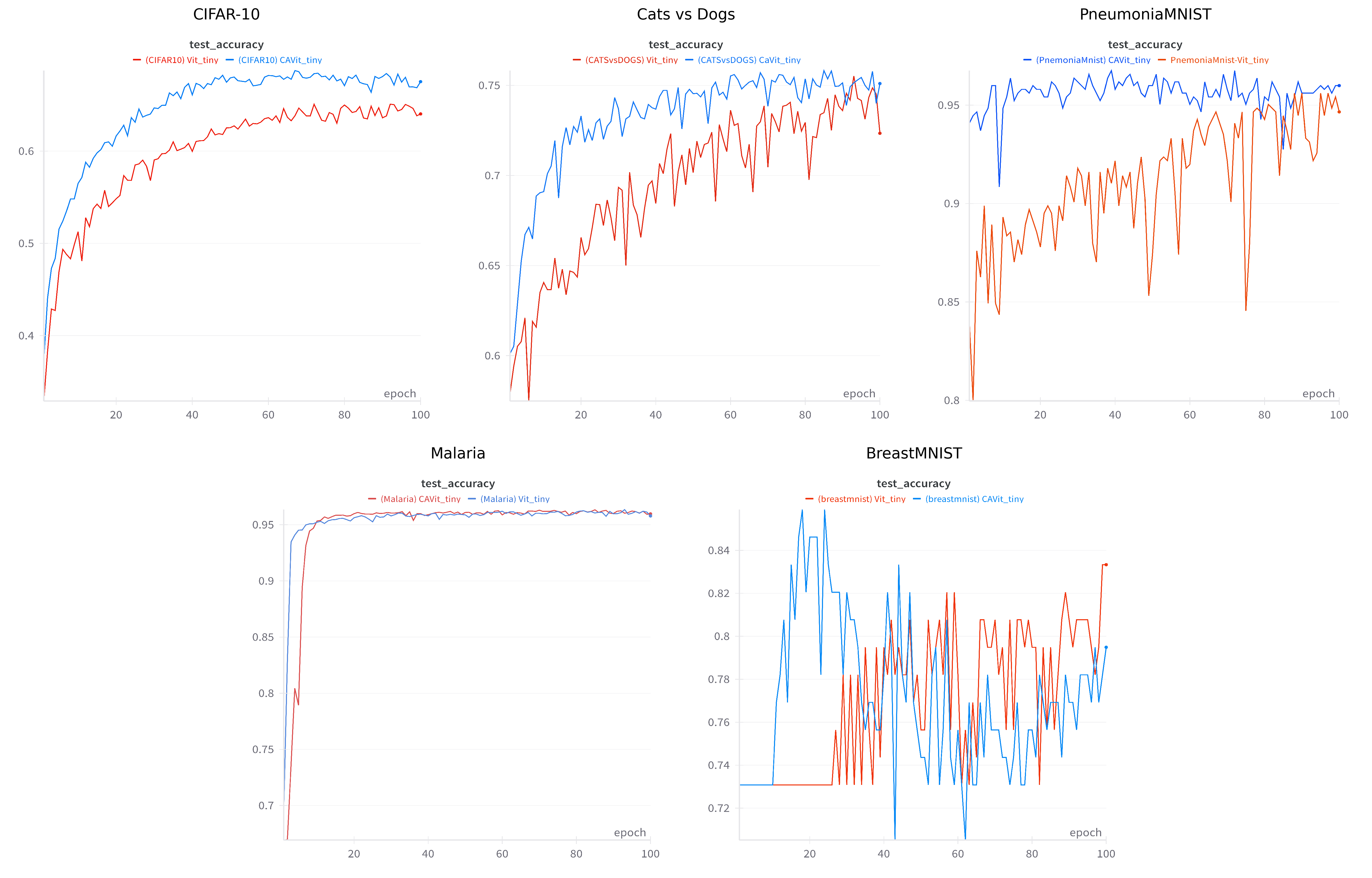}%
  }
    \centering
    \caption{\textbf{Top-1 Val Accuracy across Training Epochs.} Comparison of ViT$_{\text{tiny}}$ and CAViT$_{\text{tiny}}$ on five benchmarks. CAViT consistently shows faster convergence and better generalization on natural (CIFAR-10~\cite{cifar10}, CatsVsDogs~\cite{dogs-vs-cats}) and medical datasets (PneumoniaMNIST~\cite{yang_medmnist_2023}, Malaria~\cite{malaria_dataset}, BreastMNIST~\cite{yang_medmnist_2023}). Notably, CAViT yields consistent gains on datasets with limited resolution and inter-class variability.}
    \label{fig:accuracy_curves}
\end{figure}

\noindent
\textbf{Quantitative Performance.}
Table~\ref{tab:ca-vit-results} and Figure~\ref{fig:accuracy_curves} compare the performance of the baseline ViT$_{\text{tiny}}$ and our proposed CAViT$_{\text{tiny}}$ across five diverse datasets. CAViT achieves either comparable or superior Top-1 accuracy on all benchmarks while reducing model complexity by approximately \textbf{32\% in parameters} and \textbf{33\% in FLOPs}. Notable improvements are observed on CIFAR-10 (+3.64\%), BreastMNIST (+2.50\%), and PneumoniaMNIST (+1.14\%), indicating that CAViT generalizes better across both natural and medical image domains. Even in the Malaria dataset, where performance remains unchanged, CAViT achieves this parity with significantly fewer resources.

\begin{table}[t]
\centering
\caption{\textbf{CAViT improves accuracy across domains.} Comparison of ViT and CAViT on five diverse datasets. CAViT improves or maintains Top-1 classification accuracy in most cases, while reducing model complexity by \textbf{--32\%} in parameters ($5.75 \mapsto 3.91$M) and \textbf{--33\%} in FLOPs ($2.26 \mapsto 1.52$G).}
\vspace{0.5em}
\label{tab:ca-vit-results}
\begin{tabular}{lrrr}
\toprule
\textbf{Dataset} & $\mathrm{ViT}_{\text{tiny}}$ & $\mathrm{CAViT}_{\text{tiny}}$ & \textbf{Gain (\%)} \\
\midrule
CIFAR-10 \cite{cifar10}          & 65.09  & \textbf{68.73}  & \textbf{+3.64} \\
Malaria \cite{malaria_dataset}           & \textbf{96.35}  & 96.33  & \textbf{--0.02} \\
Cats vs Dogs \cite{dogs-vs-cats}      & 74.96  & \textbf{75.82}  & \textbf{+0.86} \\
Pneumonia \cite{yang_medmnist_2023}         & 95.61  & \textbf{96.75}  & \textbf{+1.14} \\
Breast Ultrasound \cite{yang_medmnist_2023} & 83.30  & \textbf{85.80}  & \textbf{+2.50} \\
\bottomrule
\end{tabular}
\end{table}

\vspace{0.5em}
\noindent
As shown in Figure~\ref{fig:accuracy_curves}, CAViT achieves a consistent improvement in Top-1 accuracy across most evaluated benchmarks compared to standard ViT-tiny. Particularly on CIFAR-10 and CatsVsDogs, the accuracy gains are early and stable. On medical benchmarks like PneumoniaMNIST and BreastMNIST, CAViT exhibits smoother convergence and higher peak accuracy. These results further support our architectural hypothesis that dynamic channel attention complements spatial attention to improve learning for both natural and clinical domains.

\begin{figure}[t]
  \centering
   \includegraphics[width=1.0\linewidth]{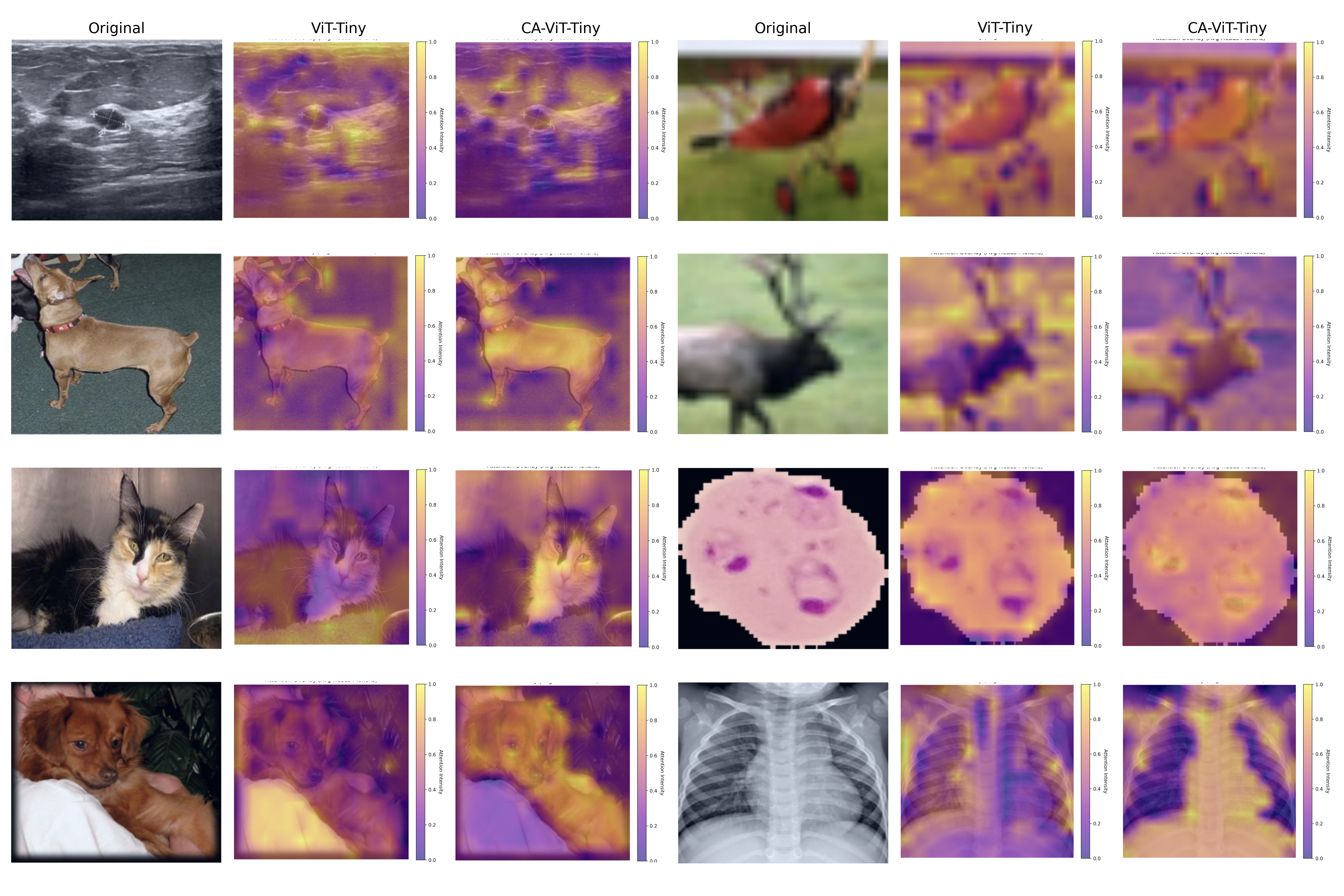}

   \caption{\textbf{Visualization of token attention:} Attention for various samples across domains including natural images, medical scans, and low-resolution categories. We use DINO-style visualization by averaging token attention maps across heads and tokens to highlight structural saliency. CAViT-Tiny captures more spatially coherent and semantically focused attention and better localizes foreground regions and anatomical features. (Best viewed in color and zoomed-in).}
   \label{fig:attnmap}
\end{figure}

\textbf{Qualitative Attention Analysis.}
To interpret the internal behavior of both models, we generate token-level attention maps using a DINO-style visualization~\cite{caron_emerging_2021}, averaging attention scores across heads and tokens. Figure~\ref{fig:attnmap} overlays these attention weights on input images. We observe that ViT$_{\text{tiny}}$ often exhibits noisy or edge-focused attention, while CAViT learns to attend to both fine-grained structures and semantically relevant regions. For instance, in medical scans (rows 3–5), CAViT highlights lesion areas more precisely, and in natural scenes, it demonstrates tighter focus on object boundaries and global form. This supports the claim that combining spatial and channel-wise attention enhances semantic reasoning. These patterns reflect the underlying benefit of channel-wise attention as it enables dynamic, content-adaptive feature fusion. Unlike static MLPs that apply fixed channel mixing across all inputs, our attention-based design recalibrates feature interactions based on global context. This improves representational expressiveness while maintaining architectural simplicity.

\subsection{Ablation Study and Architecture Insights}
\label{subsec:ablation}
To better understand the impact of different architectural decisions, we conduct a detailed ablation study summarized in Table~\ref{tab:ablation}. The variants are grouped into categories for clarity.

We first test whether spatial MHSA can be fully replaced with channel-wise SHSA. Surprisingly, this leads to a sharp performance drop across all datasets (e.g., $-2.24\%$ on CIFAR-10 and $-2.57\%$ on PneumoniaMNIST), despite a minor reduction in compute. This indicates that spatial attention is essential for capturing positional dependencies and cannot be substituted entirely by channel interactions. We further evaluate a variant of CAViT that uses multi-head self-attention in the channel branch instead of single-head SHSA. The results are slightly worse across all datasets (e.g., $-0.34\%$ on CIFAR-10), suggesting that multi-head splits in channel space may fragment semantic dependencies, and a unified global descriptor (SHSA) is more effective for channel mixing. Finally, we examine the importance of CLS token positioning by applying dimension swapping to the class token as well. This variant significantly underperforms (e.g., $-8.83\%$ on CIFAR-10 and $-5.78\%$ on PneumoniaMNIST), indicating that treating the CLS token as a channel token disrupts its intended role in aggregating spatial semantics. 

\begin{table}[t]
\centering
\caption{\textbf{Ablation study on attention configuration and CLS handling.} We isolate the effect of various components in spatial vs. channel attention.}
\setlength{\tabcolsep}{3pt}
\renewcommand{\arraystretch}{1.05}
\scriptsize
\begin{tabular}{lcccccc}
\toprule
\textbf{Ablation Variant} & \textbf{C10} & \textbf{Mal.} & \textbf{CvsD} & \textbf{Pneu.} & \textbf{Brst.} & \textbf{P/F} \\
\midrule
\multicolumn{7}{l}{\textit{Baseline - Standard ViT}} \\
Spatial\textsubscript{MHSA} $\rightarrow$ MLP  & 65.09 & \textbf{96.35} & 74.96 & 95.61 & 83.3 & 5.7M / 2.26G \\
\midrule
\multicolumn{7}{l}{\textit{Proposed - CAViT}} \\
Spatial\textsubscript{MHSA} $\rightarrow$ Channel\textsubscript{SHSA}  & \textbf{68.37} & 96.33 & \textbf{75.82} & \textbf{96.75} & \textbf{85.8} & 3.1M / 1.52G \\
\midrule
\multicolumn{7}{l}{\textit{Multi-head Channel Attention}} \\
Spatial\textsubscript{MHSA} $\rightarrow$ Channel\textsubscript{MHSA} & 64.39 & 95.43 & 74.52 & 95.37 & 82.9 & 3.1M / 1.52G \\
\midrule
\multicolumn{7}{l}{\textit{Replacing Spatial Attention}} \\
Channel\textsubscript{SHSA} $\rightarrow$ MLP & 62.85 & 92.5 & 75.13 & 94.18 & 82.73 & 5.7M / 2.26G \\
\midrule
\multicolumn{7}{l}{\textit{CLS Token Transposition}} \\
CAViT with CLS Swapped & 59.9 & 93.18 & 74.09 & 90.97 & 80.98 & 3.1M / 1.52G \\
\bottomrule
\end{tabular}
\label{tab:ablation}
\end{table}

\vspace{0.5em}
\noindent
\textbf{Discussion.}
The performance and attention patterns of CAViT reveal several key advantages. First, replacing MLP-based channel mixing with attention enables the model to dynamically reweight feature channels based on global context. Second, our dimension-swapping mechanism allows this operation to be integrated seamlessly without altering the core Transformer pipeline. The dual-path attention formulation improves representational capacity while remaining compute-efficient. Importantly, these gains are achieved \emph{without architectural deepening, extra parameters, or post-hoc regularization}, highlighting the core utility of the proposed design.

\vspace{0.5em}
\noindent
\textbf{Conclusion of Findings.}
Overall, CAViT delivers a favorable trade-off between accuracy and efficiency across a variety of domains. The model not only improves quantitative metrics but also demonstrates more interpretable and semantically grounded attention behavior. These results suggest that CAViT's proposed unified dynamic attention across spatial and feature dimensions is a promising design for Transformers in both vision and medical imaging applications.

\section{Limitations and Future Work}
\label{sec:limitations}

This work focuses on validating the core idea of combining spatial and channel-wise attention through dimension swapping within a low-resource regime. For conceptual clarity and controlled experimentation, we restrict our evaluation to ViT$_{\text{tiny}}$ and moderate-sized datasets. While this setup effectively demonstrates the advantages of our approach, it limits direct comparison with existing large-scale benchmarks. We adopt a fixed patch-based tokenizer consistent with standard ViT; however, future work may investigate how structure-aware or adaptive tokenizers can synergize with our method. We intentionally avoid training optimizations, including aggressive AugReg or distillation, and focus solely on architectural merit.  In future work, we aim to extend CAViT to multi-modal, self-supervised, and dense prediction settings.

\section{Conclusion}
\label{sec:conclusion}
We introduce CAViT, a simple and effective modification to Vision Transformers that replaces static MLP-based channel mixing with attention-based channel-wise interaction. By combining spatial and channel attention in a unified block, we improve accuracy and interpretability while reducing parameters and FLOPs. Our experiments across five diverse datasets demonstrate consistent gains over ViT$_\text{tiny}$, validating the benefit of dynamic token mixing with minimal architectural changes. This work is intended as a first step toward structured attention-based token mixing in transformers. We plan to extend our approach to larger backbones and datasets such as ImageNet. 

\begin{credits}
\subsubsection{\ackname} This publication has emanated from research conducted with the financial support of Taighde Éireann – Research Ireland under Grant number 18/CRT/6183.\footnote{For the purpose of Open Access, the author has applied a CC BY public copyright
licence to any Author Accepted Manuscript version arising from this submission.}

\subsubsection{\discintname}
The authors have no competing interests to declare that are relevant to the content of this article.
\end{credits}
%
%
%
\bibliographystyle{splncs04}
\bibliography{references}

@String(CVPR= {IEEE Conf. Comput. Vis. Pattern Recog.})

@String(ICCV= {Int. Conf. Comput. Vis.})

@String(ECCV= {Eur. Conf. Comput. Vis.})

@String(NIPS= {Adv. Neural Inform. Process. Syst.})

@String(CVPR  = {CVPR})

@String(ICCV  = {ICCV})

@String(ECCV  = {ECCV})

@String(NIPS  = {NeurIPS})

@inproceedings{ding_davit_2022,
    title = {{DaViT}: {Dual} {Attention} {Vision} {Transformers}},
    shorttitle = {{DaViT}},
    url = {https://link.springer.com/chapter/10.1007/978-3-031-20053-3_5},
    doi = {10.1007/978-3-031-20053-3_5},
    language = {en},
    urldate = {2025-04-07},
    author = {Ding, Mingyu and Xiao, Bin and Codella, Noel and Luo, Ping and Wang, Jingdong and Yuan, Lu},
    month = nov,
    year = {2022},
}

@inproceedings{ali_xcit_2021,
    title = {{XCiT}: {Cross}-{Covariance} {Image} {Transformers}},
    volume = {34},
    shorttitle = {{XCiT}},
    url = {https://proceedings.neurips.cc/paper/2021/hash/a655fbe4b8d7439994aa37ddad80de56-Abstract.html},
    urldate = {2025-04-17},
    booktitle = {Advances in {Neural} {Information} {Processing} {Systems}},
    publisher = {Curran Associates, Inc.},
    author = {Ali, Alaaeldin and Touvron, Hugo and Caron, Mathilde and Bojanowski, Piotr and Douze, Matthijs and Joulin, Armand and Laptev, Ivan and Neverova, Natalia and Synnaeve, Gabriel and Verbeek, Jakob and Jegou, Herve},
    year = {2021},
    pages = {20014--20027},
}

@misc{vaswani_attention_2023,
    title = {Attention {Is} {All} {You} {Need}},
    url = {http://arxiv.org/abs/1706.03762},
    doi = {https://doi.org/10.48550/arXiv.1706.03762},
    urldate = {2023-10-31},
    publisher = {arXiv},
    author = {Vaswani, Ashish and Shazeer, Noam and Parmar, Niki and Uszkoreit, Jakob and Jones, Llion and Gomez, Aidan N. and Kaiser, Lukasz and Polosukhin, Illia},
    month = aug,
    year = {2023},
    note = {9995 citations (Semantic Scholar/arXiv) [2023-11-07]
arXiv:1706.03762 [cs]
version: 7},
}

@inproceedings{tolstikhin_mlp-mixer_2021,
    title = {{MLP}-{Mixer}: {An} all-{MLP} {Architecture} for {Vision}},
    volume = {34},
    shorttitle = {{MLP}-{Mixer}},
    url = {https://proceedings.nips.cc/paper/2021/hash/cba0a4ee5ccd02fda0fe3f9a3e7b89fe-Abstract.html},
    urldate = {2025-03-12},
    booktitle = {Advances in {Neural} {Information} {Processing} {Systems}},
    publisher = {Curran Associates, Inc.},
    author = {Tolstikhin, Ilya O and Houlsby, Neil and Kolesnikov, Alexander and Beyer, Lucas and Zhai, Xiaohua and Unterthiner, Thomas and Yung, Jessica and Steiner, Andreas and Keysers, Daniel and Uszkoreit, Jakob and Lucic, Mario and Dosovitskiy, Alexey},
    year = {2021},
    pages = {24261--24272},
}

@misc{wu_cvt_2021,
    title = {{CvT}: {Introducing} {Convolutions} to {Vision} {Transformers}},
    shorttitle = {{CvT}},
    url = {http://arxiv.org/abs/2103.15808},
    doi = {10.48550/arXiv.2103.15808},
    urldate = {2025-03-08},
    publisher = {arXiv},
    author = {Wu, Haiping and Xiao, Bin and Codella, Noel and Liu, Mengchen and Dai, Xiyang and Yuan, Lu and Zhang, Lei},
    month = mar,
    year = {2021},
    note = {arXiv:2103.15808 [cs]},
}

@misc{trockman_patches_2022,
    title = {Patches {Are} {All} {You} {Need}?},
    url = {http://arxiv.org/abs/2201.09792},
    doi = {10.48550/arXiv.2201.09792},
    urldate = {2025-03-08},
    publisher = {arXiv},
    author = {Trockman, Asher and Kolter, J. Zico},
    month = jan,
    year = {2022},
    note = {arXiv:2201.09792 [cs]},
}

@article{khan_transformers_2022,
    title = {Transformers in {Vision}: {A} {Survey}},
    volume = {54},
    issn = {0360-0300},
    shorttitle = {Transformers in {Vision}},
    url = {https://doi.org/10.1145/3505244},
    doi = {10.1145/3505244},
    number = {10s},
    urldate = {2024-07-15},
    journal = {ACM Comput. Surv.},
    author = {Khan, Salman and Naseer, Muzammal and Hayat, Munawar and Zamir, Syed Waqas and Khan, Fahad Shahbaz and Shah, Mubarak},
    month = sep,
    year = {2022},
    pages = {200:1--200:41},
}

@article{han_survey_2023,
    title = {A {Survey} on {Vision} {Transformer}},
    volume = {45},
    issn = {1939-3539},
    url = {https://ieeexplore.ieee.org/abstract/document/9716741},
    doi = {10.1109/TPAMI.2022.3152247},
    number = {1},
    urldate = {2025-02-14},
    journal = {IEEE Transactions on Pattern Analysis and Machine Intelligence},
    author = {Han, Kai and Wang, Yunhe and Chen, Hanting and Chen, Xinghao and Guo, Jianyuan and Liu, Zhenhua and Tang, Yehui and Xiao, An and Xu, Chunjing and Xu, Yixing and Yang, Zhaohui and Zhang, Yiman and Tao, Dacheng},
    month = jan,
    year = {2023},
    note = {Conference Name: IEEE Transactions on Pattern Analysis and Machine Intelligence},
    pages = {87--110},
}

@article{shamshad_transformers_2023,
    title = {Transformers in medical imaging: {A} survey},
    volume = {88},
    issn = {1361-8415},
    shorttitle = {Transformers in medical imaging},
    url = {https://www.sciencedirect.com/science/article/pii/S1361841523000634},
    doi = {10.1016/j.media.2023.102802},
    urldate = {2023-12-03},
    journal = {Medical Image Analysis},
    author = {Shamshad, Fahad and Khan, Salman and Zamir, Syed Waqas and Khan, Muhammad Haris and Hayat, Munawar and Khan, Fahad Shahbaz and Fu, Huazhu},
    month = aug,
    year = {2023},
    pages = {102802},
}

@misc{bordes_introduction_2024,
    title = {An {Introduction} to {Vision}-{Language} {Modeling}},
    url = {http://arxiv.org/abs/2405.17247},
    doi = {10.48550/arXiv.2405.17247},
    urldate = {2025-04-17},
    publisher = {arXiv},
    author = {Bordes, Florian and Pang, Richard Yuanzhe and Ajay, Anurag and Li, Alexander C. and Bardes, Adrien and Petryk, Suzanne and Mañas, Oscar and Lin, Zhiqiu and Mahmoud, Anas and Jayaraman, Bargav and Ibrahim, Mark and Hall, Melissa and Xiong, Yunyang and Lebensold, Jonathan and Ross, Candace and Jayakumar, Srihari and Guo, Chuan and Bouchacourt, Diane and Al-Tahan, Haider and Padthe, Karthik and Sharma, Vasu and Xu, Hu and Tan, Xiaoqing Ellen and Richards, Megan and Lavoie, Samuel and Astolfi, Pietro and Hemmat, Reyhane Askari and Chen, Jun and Tirumala, Kushal and Assouel, Rim and Moayeri, Mazda and Talattof, Arjang and Chaudhuri, Kamalika and Liu, Zechun and Chen, Xilun and Garrido, Quentin and Ullrich, Karen and Agrawal, Aishwarya and Saenko, Kate and Celikyilmaz, Asli and Chandra, Vikas},
    month = may,
    year = {2024},
    note = {arXiv:2405.17247 [cs]},
}

@misc{zhou_image_2024,
    title = {Image {Segmentation} in {Foundation} {Model} {Era}: {A} {Survey}},
    shorttitle = {Image {Segmentation} in {Foundation} {Model} {Era}},
    url = {http://arxiv.org/abs/2408.12957},
    doi = {10.48550/arXiv.2408.12957},
    urldate = {2025-04-17},
    publisher = {arXiv},
    author = {Zhou, Tianfei and Xia, Wang and Zhang, Fei and Chang, Boyu and Wang, Wenguan and Yuan, Ye and Konukoglu, Ender and Cremers, Daniel},
    month = nov,
    year = {2024},
    note = {arXiv:2408.12957 [cs]},
}

@inproceedings{kahatapitiya_swat_2023,
    title = {{SWAT}: {Spatial} {Structure} {Within} and {Among} {Tokens}},
    volume = {2},
    shorttitle = {{SWAT}},
    url = {https://www.ijcai.org/proceedings/2023/106},
    doi = {10.24963/ijcai.2023/106},
    language = {en},
    urldate = {2025-02-13},
    author = {Kahatapitiya, Kumara and Ryoo, Michael S.},
    month = aug,
    year = {2023},
    note = {ISSN: 1045-0823},
    pages = {956--964},
}

@inproceedings{liu_swin_2021,
    address = {Montreal, QC, Canada},
    title = {Swin {Transformer}: {Hierarchical} {Vision} {Transformer} using {Shifted} {Windows}},
    copyright = {https://doi.org/10.15223/policy-029},
    isbn = {978-1-66542-812-5},
    shorttitle = {Swin {Transformer}},
    url = {https://ieeexplore.ieee.org/document/9710580/},
    doi = {10.1109/ICCV48922.2021.00986},
    language = {en},
    urldate = {2025-04-17},
    booktitle = {2021 {IEEE}/{CVF} {International} {Conference} on {Computer} {Vision} ({ICCV})},
    publisher = {IEEE},
    author = {Liu, Ze and Lin, Yutong and Cao, Yue and Hu, Han and Wei, Yixuan and Zhang, Zheng and Lin, Stephen and Guo, Baining},
    month = oct,
    year = {2021},
    pages = {9992--10002},
}

@inproceedings{caron_emerging_2021,
    address = {Montreal, QC, Canada},
    title = {Emerging {Properties} in {Self}-{Supervised} {Vision} {Transformers}},
    copyright = {https://doi.org/10.15223/policy-029},
    isbn = {978-1-66542-812-5},
    url = {https://ieeexplore.ieee.org/document/9709990/},
    doi = {10.1109/ICCV48922.2021.00951},
    language = {en},
    urldate = {2025-02-28},
    booktitle = {2021 {IEEE}/{CVF} {International} {Conference} on {Computer} {Vision} ({ICCV})},
    publisher = {IEEE},
    author = {Caron, Mathilde and Touvron, Hugo and Misra, Ishan and Jegou, Herve and Mairal, Julien and Bojanowski, Piotr and Joulin, Armand},
    month = oct,
    year = {2021},
    pages = {9630--9640},
}

@inproceedings{touvron_training_2021,
    title = {Training data-efficient image transformers \& distillation through attention},
    url = {https://proceedings.mlr.press/v139/touvron21a.html},
    language = {en},
    urldate = {2025-04-17},
    booktitle = {Proceedings of the 38th {International} {Conference} on {Machine} {Learning}},
    publisher = {PMLR},
    author = {Touvron, Hugo and Cord, Matthieu and Douze, Matthijs and Massa, Francisco and Sablayrolles, Alexandre and Jegou, Herve},
    month = jul,
    year = {2021},
    note = {ISSN: 2640-3498},
    pages = {10347--10357},
}

@article{cifar10,
  title={Learning multiple layers of features from tiny images},
  author={Krizhevsky, Alex and others},
  year={2009}
}

@article{yang_medmnist_2023,
    title = {{MedMNIST} v2 - {A} large-scale lightweight benchmark for {2D} and {3D} biomedical image classification},
    volume = {10},
    copyright = {2023 The Author(s)},
    issn = {2052-4463},
    url = {https://www.nature.com/articles/s41597-022-01721-8},
    doi = {10.1038/s41597-022-01721-8},
    language = {en},
    number = {1},
    urldate = {2025-04-18},
    journal = {Scientific Data},
    author = {Yang, Jiancheng and Shi, Rui and Wei, Donglai and Liu, Zequan and Zhao, Lin and Ke, Bilian and Pfister, Hanspeter and Ni, Bingbing},
    month = jan,
    year = {2023},
    note = {Publisher: Nature Publishing Group},
    pages = {41},
}

@InProceedings{Hu_2018_CVPR,
author = {Hu, Jie and Shen, Li and Sun, Gang},
title = {Squeeze-and-Excitation Networks},
booktitle = {Proceedings of the IEEE Conference on Computer Vision and Pattern Recognition (CVPR)},
month = {June},
year = {2018}
}

@InProceedings{Woo_2018_ECCV,
author = {Woo, Sanghyun and Park, Jongchan and Lee, Joon-Young and Kweon, In So},
title = {CBAM: Convolutional Block Attention Module},
booktitle = {Proceedings of the European Conference on Computer Vision (ECCV)},
month = {September},
year = {2018}
}

@inproceedings{liu_pay_2021,
    address = {Red Hook, NY, USA},
    series = {{NIPS} '21},
    title = {Pay attention to {MLPs}},
    isbn = {978-1-71384-539-3},
    urldate = {2025-04-18},
    booktitle = {Proceedings of the 35th {International} {Conference} on {Neural} {Information} {Processing} {Systems}},
    publisher = {Curran Associates Inc.},
    author = {Liu, Hanxiao and Dai, Zihang and So, David R. and Le, Quoc V.},
    month = dec,
    year = {2021},
    pages = {9204--9215},
}

@inproceedings{yuan_tokens--token_2021,
    address = {Montreal, QC, Canada},
    title = {Tokens-to-{Token} {ViT}: {Training} {Vision} {Transformers} from {Scratch} on {ImageNet}},
    copyright = {https://doi.org/10.15223/policy-029},
    isbn = {978-1-66542-812-5},
    shorttitle = {Tokens-to-{Token} {ViT}},
    url = {https://ieeexplore.ieee.org/document/9710747/},
    doi = {10.1109/ICCV48922.2021.00060},
    language = {en},
    urldate = {2025-03-26},
    booktitle = {2021 {IEEE}/{CVF} {International} {Conference} on {Computer} {Vision} ({ICCV})},
    publisher = {IEEE},
    author = {Yuan, Li and Chen, Yunpeng and Wang, Tao and Yu, Weihao and Shi, Yujun and Jiang, Zihang and Tay, Francis E. H. and Feng, Jiashi and Yan, Shuicheng},
    month = oct,
    year = {2021},
    pages = {538--547},
}

@software{Wightman_PyTorch_Image_Models,
author = {Wightman, Ross},
doi = {10.5281/zenodo.4414861},
license = {Apache 2.0},
title = {{PyTorch Image Models}},
url = {https://github.com/huggingface/pytorch-image-models},
version = {1.0.11}
}

@misc{malaria_dataset,
  author       = {I. Arunava},
  title        = {Cell Images for Detecting Malaria},
  year         = {2018},
  howpublished = {\url{https://www.kaggle.com/datasets/iarunava/cell-images-for-detecting-malaria}},
  note         = {Accessed: 2025-04-18}
}

@misc{dogs-vs-cats,
    author = {Will Cukierski},
    title = {Dogs vs. Cats},
    year = {2013},
    howpublished = {\url{https://kaggle.com/competitions/dogs-vs-cats}},
    note = {Kaggle}
}

@inproceedings{
newdosovitskiy2021an,
title={An Image is Worth 16x16 Words: Transformers for Image Recognition at Scale},
author={Alexey Dosovitskiy and Lucas Beyer and Alexander Kolesnikov and Dirk Weissenborn and Xiaohua Zhai and Thomas Unterthiner and Mostafa Dehghani and Matthias Minderer and Georg Heigold and Sylvain Gelly and Jakob Uszkoreit and Neil Houlsby},
booktitle={International Conference on Learning Representations},
year={2021},
url={https://openreview.net/forum?id=YicbFdNTTy}
}

@inproceedings{zheng_lightweight_2023,
    title = {Lightweight {Vision} {Transformer} with {Spatial} and {Channel} {Enhanced} {Self}-{Attention}},
    booktitle = {2023 {IEEE}/{CVF} {International} {Conference} on {Computer} {Vision} {Workshops} ({ICCVW})},
    author = {Zheng, Jiahao and Yang, Longqi and Li, Yiying and Yang},
    month = oct,
    year = {2023},
    pages = {1484--1488},
}

@inproceedings{Fu_2019_CVPR,
  author    = {Fu, Jun and others},
  title     = {Dual Attention Network for Scene Segmentation},
  booktitle = {Proceedings of the IEEE/CVF Conference on Computer Vision and Pattern Recognition (CVPR)},
  year      = {2019},
  month     = {June}
}
%




\end{document}